%% file: main.tex
\documentclass[10pt,twocolumn,letterpaper]{article}
\pdfoutput=1
\usepackage[pagenumbers]{cvpr} 

\input{preamble}

\definecolor{cvprblue}{rgb}{0.21,1,0}
\usepackage[pagebackref,breaklinks,colorlinks,citecolor=cvprblue]{hyperref}

\algnewcommand\algorithmicinput{\textbf{Input:}}
\algnewcommand\algorithmicoutput{\textbf{Output:}}
\algnewcommand\algorithmicinit{\textbf{Initialize:}}
\algnewcommand\Input{\item[\algorithmicinput]}
\algnewcommand\Output{\item[\algorithmicoutput]}
\algnewcommand\Initialize{\item[\algorithmicinit]}

\def\paperID{6251} 
\def\confName{CVPR}
\def\confYear{2024}

\title{Class-Adaptive Sampling Policy for Efficient Continual Learning}

\author{Hossein Rezaei\\
School of Computer Engineering\\
Iran University of Science and Technology (IUST)\\
\href{mailto:hossein_rezaei@comp.iust.ac.ir}{\textcolor{black}{\tt\small hossein\_rezaei@comp.iust.ac.ir}}
\and
Mohammad Sabokrou\\
Okinawa Institute of Science and Technology (OIST)\\
{\tt\small mohammad.sabokrou@oist.jp}
}

\begin{document}
\maketitle
\input{sec/0_abstract}    
\input{sec/1_intro}

\input{sec/3_Preliminaries}

\input{sec/4_method}

\input{sec/5_Contribution}

\input{sec/6_Experiments}

\input{sec/7_Conclusion}

{
    \small
    \bibliographystyle{ieeenat_fullname}
    \bibliography{main}
}

\input{sec/8_Supplementary}


\end{document}

%% file: preamble.tex
\usepackage[dvipsnames]{xcolor}

\usepackage{multirow}
\usepackage{arydshln}
\usepackage{amsmath}

\usepackage{algorithm}
\usepackage{algpseudocode}

%% file: sec/0_abstract.tex
\begin{abstract}
 Continual learning (CL) aims to acquire new knowledge while preserving information from previous experiences without forgetting. Though buffer-based methods (i.e., retaining samples from previous tasks) have achieved acceptable performance, determining how to allocate the buffer remains a critical challenge. Most recent research focuses on refining these methods but often fails to sufficiently consider the varying influence of samples on the learning process, and frequently overlooks the complexity of the classes/concepts being learned. Generally, these methods do not directly take into account the contribution of individual classes. However, our investigation indicates that more challenging classes necessitate preserving a larger number of samples compared to less challenging ones. To address this issue, we propose a novel method and policy named 'Class-Adaptive Sampling Policy' (CASP), which dynamically allocates storage space within the buffer. By utilizing concepts of class contribution and difficulty, CASP adaptively manages buffer space, allowing certain classes to occupy a larger portion of the buffer while reducing storage for others. This approach significantly improves the efficiency of knowledge retention and utilization.
CASP provides a versatile solution to boost the performance and efficiency of CL. It meets the demand for dynamic buffer allocation, accommodating the varying contributions of different classes and their learning complexities over time.\footnote{Codes are available at: \href{https://github.com/hossein-rezaei624/CASP}{https://github.com/hossein-rezaei624/CASP}}
\end{abstract}

%% file: sec/1_intro.tex
\section{Introduction}
\label{sec:intro}

\begin{figure}[t] 
  \centering
  \includegraphics[width=0.47\textwidth]{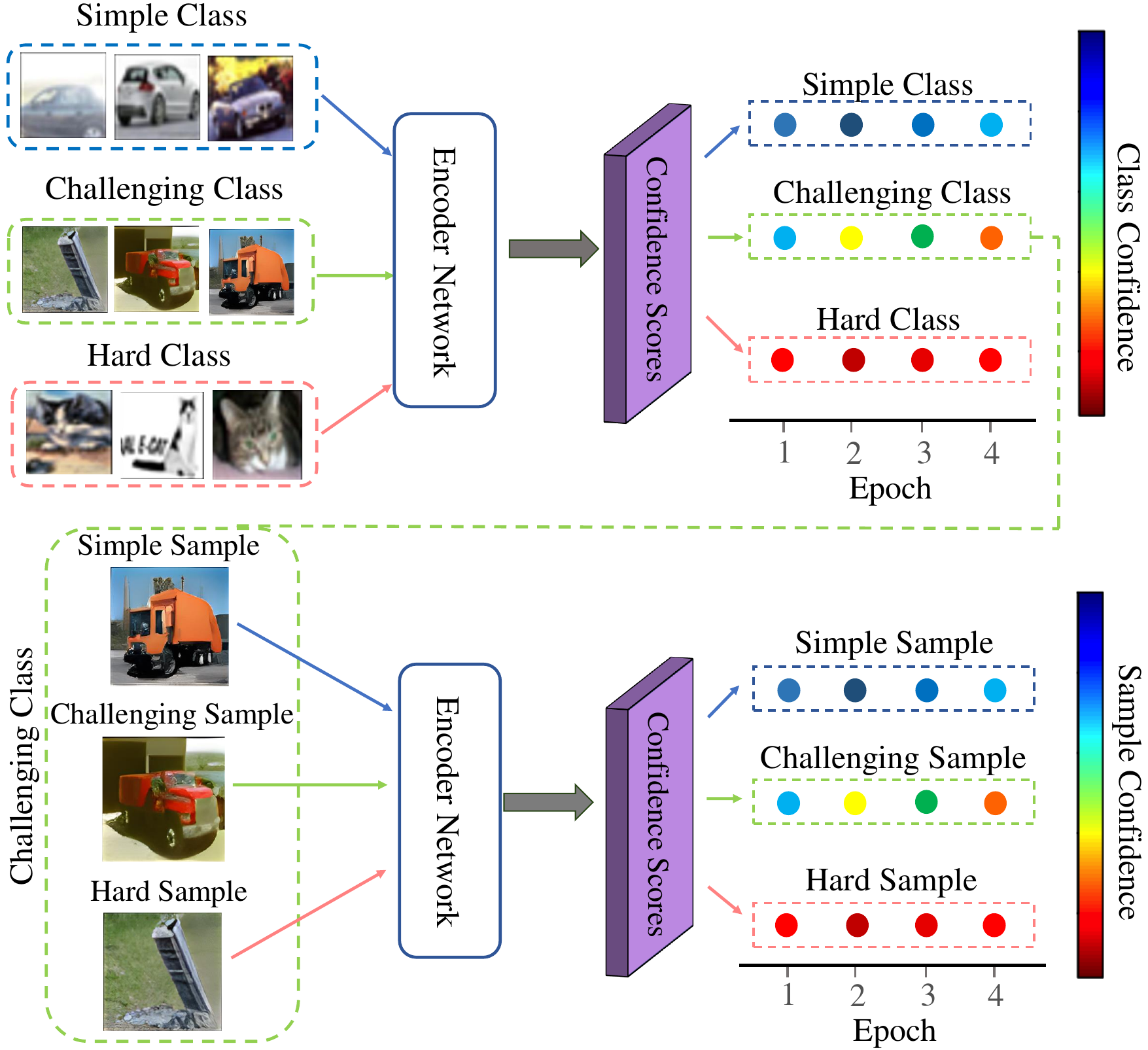}
  \caption{Each circle symbolizes the model's confidence level for a specific class/sample during a given epoch. A blue hue signifies high confidence, while red indicates low confidence. The model uses a ResNet18 Encoder as its backbone. Our method focuses more intently on the 'truck' class due to its learning complexity. Moreover, the CASP approach emphasizes challenging samples within each class, as they offer more representative insights.}
  \label{fig:Framework}
\end{figure}

In the real world, it is a common scenario where a machine learning model, during the testing phase, encounters new tasks or classes that it needs to learn without having access to the training data from its previous experiences. In such cases, these models often significantly forget their previous experiences. To address this challenge, Continual Learning (CL) has recently attracted substantial attention from researchers. Although several approaches have been proposed, the main challenge persists in balancing the learning of new concepts (i.e., plasticity) while preserving existing knowledge (i.e., stability) \cite{kim2023achieving, kim2023stability}.

Regularization techniques ~\cite{kirkpatrick2017overcoming, douillard2020podnet, sun2023regularizing},  architecture based ~\cite{mallya2018packnet, liu2021adaptive, hu2023dense} and buffer based~\cite{chaudhry2019tiny, NEURIPS2019_15825aee, aljundi2019gradient, shim2021online, mai2021supervised, gu2022not, lin2023pcr,de2021continual} are three major approaches for CL. The literature review shows that retaining and utilizing samples from previous classes or tasks in a buffer (i.e., buffer-based methods) delivers a better performance  than two other solutions. 

Generally, buffer-based methods utilize a small memory buffer populated with samples from previously learned classes/tasks to avoid catastrophic forgetting, and the model sees some samples from previous tasks/classes alongside new tasks. During the learning process, as each batch arrives, they retrieve a select number of samples from the buffer to relearn earlier classes in tandem with the new ones. Subsequently, the buffer is updated with samples from the current batch. Most buffer-based methods, such as {ER} ~\cite{chaudhry2019tiny}, {MIR}~\cite{NEURIPS2019_15825aee}, {SCR}~\cite{mai2021supervised}, {DVC}~\cite{gu2022not}, and {PCR}~\cite{lin2023pcr}, treat all samples uniformly when updating the buffer, thereby overlooking the varying contributions of different classes to this process.

In the literature, several methods have been proposed for sample selection to be retained in the buffer. For instance, GSS \cite{aljundi2019gradient}, GMED\cite{jin2021gradient}, OCS\cite{yoon2021online}, and GCR\cite{tiwari2022gcr} utilize the gradients of samples to select the more informative ones. Meanwhile, ASER\cite{shim2021online}, RAR\cite{kumari2022retrospective,yao2023continual},  and \cite{hu2022curiosity} propose sampling from areas near the decision boundary. CoPE \cite{de2021continual} employs class prototype samples. Furthermore, \cite{wang2022improving} uses Distributionally Robust Optimization (DRO) to encourage the buffer’s distribution to become more complex. However, this method focuses on sample selection, a notable weakness is their lack of consideration for the contribution of individual classes/concepts to the buffer's composition (In the supplementary material, we comprehensively and in detail discussed the related works.). 


\begin{figure}[t]
  \centering
  \includegraphics[width=0.27\textwidth]{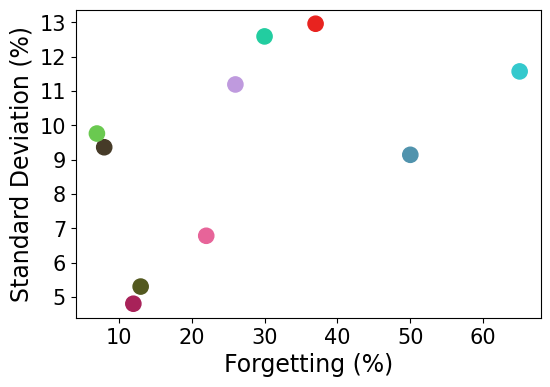}
  \caption{We analyzed the first 10 classes of Split CIFAR100 as the initial task and assessed the forgetting of each class during the continual learning of 10 tasks. Additionally, we trained these first 10 classes in a separate model over \(E\) epochs and measured the standard deviation of each class's confidence scores. This analysis revealed a moderate Pearson correlation, with a coefficient of 50.66\(\%\), between them. Each dot represents a distinct class.}
  \label{fig:Correlation}
\end{figure}

In line with our earlier discussion, we recognize that certain classes or concepts pose more challenges than others, just as individual samples vary in their level of difficulty. Previous research reveals that not all samples are equally informative; some contribute substantially more to the learning process than others. This is highlighted in studies focusing on data influence estimation, such as those by ~\cite{toneva2018empirical, swayamdipta2020dataset}, along with related research by ~\cite{motamedi2021data, mazumder2022dataperf}. These studies underscore the varying value of different samples. Consequently, this leads us to question the effectiveness of a uniform approach in sample selection for storage. We propose that selectively retaining samples that more effectively represent previous classes, rather than choosing them randomly, could significantly improve performance and reduce the likelihood of forgetting. 



In a nutshell, this paper aims to address the following questions: 
\textit{(1) Does the choice of sampling approaches have an impact on reducing forgetting? (2) Can increased attention to the challenging classes/concepts and allocating more space for them in the buffer improve performance?}

The first question has not been enough attended while the second one has remained unexplored. In a brief response to both questions, our results demonstrate that the answer is affirmative. To address these issues, we propose an innovative approach called the Class-Adaptive Sampling Policy (CASP) (See Fig.~\ref{fig:Framework}), which formulates a strategy for selecting samples and determining which classes require greater attention.

In the training of current classes during ongoing tasks, CASP considers the model's confidence in each sample across different epochs. This serves as a measure to assess the suitability of samples for retention at the end of training and for buffer allocation. Typically, samples that the model consistently labels with high confidence are deemed simple, whereas those labeled with very low confidence are viewed as hard (outliers). However, there is a distinct category of samples that pose a challenge to the model, indicated by significant fluctuations in the model's confidence during training. These samples exhibit a high standard deviation of confidence across epochs, highlighting their challenging nature. CASP prioritizes these challenging samples for retention because they represent the most significant obstacles encountered by the model. To assess the difficulty of a class and determine its share of the buffer, CASP calculates the average confidence of samples within each class at every epoch, defining this metric as 'class confidence.' The standard deviation of class confidence across epochs serves as a gauge of the class's difficulty.\footnote{Classes are also categorized based on confidence levels: 'challenging' for significant fluctuations, 'simple' for consistently high, and 'hard' for persistently low.} CASP conducts this analysis for all classes, followed by normalization of the results. This normalization enables us to proportionately allocate buffer space to each class, aligning it with the evaluated level of challenge.  Intuitively, the challenge level of classes, indicated by their standard deviation scores, directly correlates with their forgettability — how prone a model is to forget them. This concept of forgettability refers to the model's tendency to lose grasp of specific classes over time. Fig.~\ref{fig:Correlation} illustrates this phenomenon, where each dot represents a class, visually depicting the relationship between class difficulty and forgettability

Our proposed methods can function as a plug-in for all buffer-based approaches, significantly boosting their effectiveness. Owing to the data-centric strategy utilized in our approach, the model demonstrates enhanced generalizability over prior methods. This improvement is especially notable in our assessment of out-of-distribution (OOD) generalization, as elaborated in the experiments section \ref{Ablation}. 



%% file: sec/3_Preliminaries.tex
\section{Preliminaries} 
\label{sec:Preliminaries}




Consider a task, denoted by \( \mathcal{D}_t \), which comprises classes \( \mathcal{C}_j \), and a specified space \( \mathcal{M}_t \) in the buffer \( \mathbf{B}_t \), observed by a model \( f_{\theta_{t}} \) at a specific time \( t \). \( \mathcal{D}_t \) is comprised of pairs \( (X_i, Y_i) \), where \( X_i \) represents the input data and \( Y_i \) is its corresponding label or target. As the model progresses to a new task, \( \mathcal{D}_{\text{new}} \), at time \( t+1 \), the goal shifts to updating the model, now denoted as \( f_{\theta_{t+1}} \). This update is informed by the new task data \( \mathcal{D}_{\text{new}} \) and a selected portion of \( \mathcal{D}_{\leq{t}} \) retained in the buffer \( \mathbf{B}_t \). The primary objective of this update is to minimize the loss function, \( \mathcal{L} \), across the combined datasets of \( D_{t+1} \) and the contents of the buffer \( \mathbf{B}_t \).

\begin{equation}
f_{\theta_{t+1}} = \arg\min_{\theta} [\mathcal{L}(D_{t+1}, \mathbf{B}_t, f_{\theta_t})]
\end{equation}

In the realm of task incremental learning, particularly when dealing with previous tasks characterized by unique classes, the predominant focus of existing research has been narrowly confined to assessing the contribution of samples, while largely neglecting the vital role played by individual classes. Our study, however, underscores the critical necessity of formulating a strategy that prioritizes and meticulously addresses the complexities and nuances inherent in the more challenging classes. In this approach, the samples in \(\mathbf{B}\) are dynamically updated after each task, according to a class selection policy. This policy, expressed as \(\mathcal{M}_t = \pi_{cs}(\mathcal{V}((X_i, Y_i), \mathcal{C}_j, \mathcal{D}_t, \hat{f}_{\hat{\theta}}), \pi_{ss})\), involves \(\mathcal{V}(.)\), a function that assesses the vulnerability level of class \(\mathcal{C}_j\). Additionally, \(\pi_{ss}\) represents the sample selection strategy for class \(\mathcal{C}_j\), defined as \(\pi_{ss}((X_i, Y_i), \mathcal{D}_t, \hat{f}_{\hat{\theta}})\), where \(\hat{f}_{\hat{\theta}}\) is a surrogate model designed specifically for CASP, distinct from the main CL model. This tailored strategy determines which samples are pivotal for each class in the task and facilitates the transfer of knowledge to the model.

%% file: sec/4_method.tex
\section{Proposed Method}
\label{sec:Proposed Method}
In our proposed methodology, we emphasize the need for improved focus on memory space allocation in the buffer for retained samples, addressing aspects previously overlooked in CL. We introduce the CASP method as a versatile plug-and-play solution to enhance existing CL methods, highlighting the lack of generalization in current CL approaches. 

Firstly, we introduce the policy \(\pi_{cs}\). This policy is instrumental in identifying which classes necessitate a heightened level of attention to mitigate the risk of forgetting. By analytically estimating the fluctuations in the class's confidence during the learning process, we define a ‘vulnerability score’, \(\mathcal{V}(.)\), for each class \(\mathcal{C}_{j=0}^{j=k}\) post the learning phase of \(f_{\theta_{t}}\). This scoring system enables us to judiciously allocate more buffer space to classes that are deemed more challenging, while concurrently allocating less space to those that are less challenging.

Parallelly, we employ a similar conceptual framework to formulate the policy \(\pi_{ss}\): a strategic approach for selecting specific samples from the $j^{th}$ class in the $t^{th}$ task post the training of \(f_{\theta_{t}}\), aiming to retain the most crucial samples rather than relying on a random selection policy. At the conclusion of the training \(f_{\theta_{t}}\) encompassing \(K\) distinct classes, we update \(\mathbf{B}_t\) by retaining those samples that are deemed most likely to be forgotten. This is meticulously determined by tracking the fluctuations in the model’s confidence levels across various samples during the different epochs of the training task \(\mathcal{D}_t\).

Intriguingly, our findings reveal that the most forgettable/challenging samples, barring outliers, tend to be the most representative of the entire data distribution. 

In the following, we delve into the specifics of our newly introduced strategies for updating the memory buffer. Also, The integration of our method with the Experience Replay (ER) method is exemplified in Algorithm \ref{alg:er_method}.


\begin{algorithm}
\scalebox{0.85}{
\begin{minipage}{1.0\linewidth}
\caption{Integration of CASP with the ER Method}
\label{alg:er_method}
\begin{algorithmic}[1]
\Input \(\mathcal{D}\), \(\mathbf{B}\), \(\hat{f}_{\hat{\theta}}\), \(\theta\), \( E\)
\Output \(\mathbf{B}\), \(\theta\)
\Initialize \(\mathbf{B} \leftarrow \{\}\); \(b \leftarrow 0\); Parameters \(\theta\) 
\For{\( t = 0 \) to \( T \)}
    \State Initialize \(\hat{\theta}\) 
    \For{\(Batch_b \sim \mathcal{D}_t\)}
        \State \( Batch_q \leftarrow \) RandomRetrieval(\(\mathbf{B}_{t}\))
        \State \( \theta \leftarrow \) SGD(\( Batch_b \cup Batch_q, \theta\))
        \State \( \mathbf{B}_{t} \leftarrow \) RandomUpdate(\( Batch_b, \mathbf{B}_{t} \))
    \EndFor
    \State \(\mathcal{M}_t \leftarrow\) CASP\((\mathcal{D}_t, \hat{f}_{\hat{\theta}}, E)\) \Comment{The space given to task \(\mathcal{D}_t\) by RandomUpdate(.) from buffer \(\mathbf{B}_{t}\) is called \(\mathcal{M}_t\).}
\EndFor
\end{algorithmic}
\end{minipage}
}
\end{algorithm}

\subsection{$\pi_{cs}$:Class Strategy}
When we encounter a new task, denoted as \(\mathcal{D}_t\), we train a surrogate model, represented as \(\hat{f}_{\hat{\theta}}\), from scratch over \(E\) epochs and calculate the expected target SoftMax scores for each class. Upon analyzing this value over multiple epochs, we observed that it continuously fluctuates for classes that tend to be forgotten. As a result, for classes with higher variance, we allocate more space in the buffer. 
The formulations are as follows:

\begin{equation}
\Gamma{(\mathcal{C}_{j}, e)} = \mathbb{E}_{X \sim \mathcal{C}_j}\left[\mathcal{P}_{\hat{\theta}_{e}} (Y | X)\right]
\end{equation}

Where \(\Gamma{(\mathcal{C}_{j}, e)}\) represents the confidence level of the \(j^{th}\) class at epoch \(e\), and \( \mathcal{P}_{\hat{\theta}_{e}}(Y | X) \) indicates the probability of the target class given the input, with the model parameters at epoch \(e\). The expectation is taken over the distribution of samples \(X\) in class \(j\).

In the next step, we assess the vulnerability of each class across \( E \) epochs by computing the standard deviation, which is outlined as follows:

\begin{equation}
\label{eq:mean_cf}
\bar{\Gamma}(\mathcal{C}_{j}) = \mathbb{E}_{e \sim \{0, 1, \ldots, E\}}[\Gamma(\mathcal{C}_{j}, e)]
\end{equation}

\begin{equation}
\label{eq:var_cf}
\mathcal{V}(\mathcal{C}_{j}) =  \sqrt{\mathbb{E}_{e \sim \{0, 1, \ldots, E\}}\left[\left(\Gamma(\mathcal{C}_{j}, e) - \bar{\Gamma}(\mathcal{C}_{j})\right)^2\right]}
\end{equation}

Here, \(\mathcal{V}(\mathcal{C}_{j})\) represents the vulnerability (variability in confidence levels) of class \(j\) across epochs. The Equation~\ref{eq:var_cf} yields the standard deviation by computing the square root of the variance in confidence values over \(E\) epochs, while the Equation~\ref{eq:mean_cf} calculates the mean confidence level of class \(j\) across all epochs.

In the subsequent step, based on the size of \(\mathcal{M}_t\), we determine the number of samples to be allocated for each class \(\mathcal{C}_j\) within the current task, \(\mathcal{D}_t\). This is described as follows:

\begin{equation}
\mathcal{S}_j = \frac{\mathcal{V}(\mathcal{C}_{j})}{\sum_{j=0}^K \mathcal{V}(\mathcal{C}_{j})}\times \mathcal{M}_t
\end{equation}
Where \( \frac{\mathcal{V}(\mathcal{C}_{j})}{\sum_{j=0}^K \mathcal{V}(\mathcal{C}_{j})}\) denotes the normalized \(\mathcal{V}(\mathcal{C}_{j})\).

\subsection{$\pi_{ss}$:Sample Strategy}

Drawing inspiration from the works of ~\cite{toneva2018empirical} and ~\cite{swayamdipta2020dataset}, we propose a novel method for identifying more meaningful samples within each class. Our approach categorizes samples into three distinct groups based on their behavior during the learning process at epoch E. The first group consists of samples that exhibit low SoftMax values, indicating a lack of confidence in the prediction. The second group comprises samples that demonstrate high SoftMax values, reflecting a high degree of confidence. Finally, the third group contains samples characterized by a high variance in SoftMax values, signifying fluctuation in the model's certainty across the learning epochs. This classification allows for a nuanced understanding of the model's learning dynamics and facilitates targeted improvements in its performance. Upon analysis, we have determined that the samples within the third group—those exhibiting high variance in SoftMax values—possess the most distinctive features of their respective classes. Consequently, we focus on leveraging these samples for each class. Our methodological formulations are as follows:

\begin{equation}
\bar{\Gamma}({X}_{i}) = \mathbb{E}_{e \sim \{0, 1, \ldots, E\}}\left[\mathcal{P}_{\hat{\theta}_{e}} (Y | {X}_{i})\right]
\end{equation}

Here, \(\bar{\Gamma}({X}_{i})\) denotes the average confidence level for the \(i^{th}\) sample after completing learning at epoch \(E\). The term \( \mathcal{P}_{\hat{\theta}_{e}}(Y | {X}_{i}) \) represents the conditional probability of the target class \(Y\) given the input \({X}_{i}\), parameterized by the model at epoch \(e\). This expectation is computed over all epochs up to E for the task \(\mathcal{D}_t\).

Following this, we determine the standard deviation of each sample within the task \(\mathcal{D}_t\):

\begin{equation}
\mathcal{V}({X}_{i}) =  \sqrt{\mathbb{E}_{e \sim \{0, 1, \ldots, E\}}\left[\left(\mathcal{P}_{\hat{\theta}_{e}} (Y | {X}_{i}) - \bar{\Gamma}({X}_{i})\right)^2\right]}
\end{equation}

In this expression, \(\mathcal{V}({X}_{i})\) signifies the vulnerability of the \(i^{th}\) sample across the learning epochs up to \(E\) in task \(\mathcal{D}_t\), capturing the fluctuations in the model's confidence for that sample.

%% file: sec/5_Contribution.tex
\section{Contribution of different samples/classes}
\label{sec:Contribution}

\begin{figure}[t]
  \includegraphics[width=0.4\textwidth]{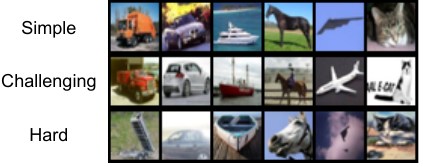}
  \caption{The first row displays simple examples from four classes of CIFAR10 ~\cite{krizhevsky2009learning} (see supplementary material for other classes). In contrast, the second row presents challenging examples, and the third row features the hard samples for these classes. The simple samples are easily identifiable, showcasing prominent class attributes or centrally-placed objects, such as a horse on grass. Challenging examples, however, display more ambiguous traits like an airplane against a gray backdrop, deviating from the typical learning patterns associated with their class. Hard samples are characterized by unusual or rare features, for instance, a horse where only the head is visible. These hard examples often function as outliers, diverging significantly from the standard distribution of samples.}
  \label{fig:SelectedSamples}
\end{figure}

We train the CIFAR10 dataset over a few epochs in an offline mode. During each epoch, we measure the confidence scores for each class and sample. Based on these scores, we categorize the samples/classes into three groups: those with high average confidence across the epochs are deemed simple, those with low average confidence are labeled as hard, and those with high standard deviation in confidence scores are considered challenging. 

\begin{figure}[t]
  \centering
  \includegraphics[width=0.47\textwidth]{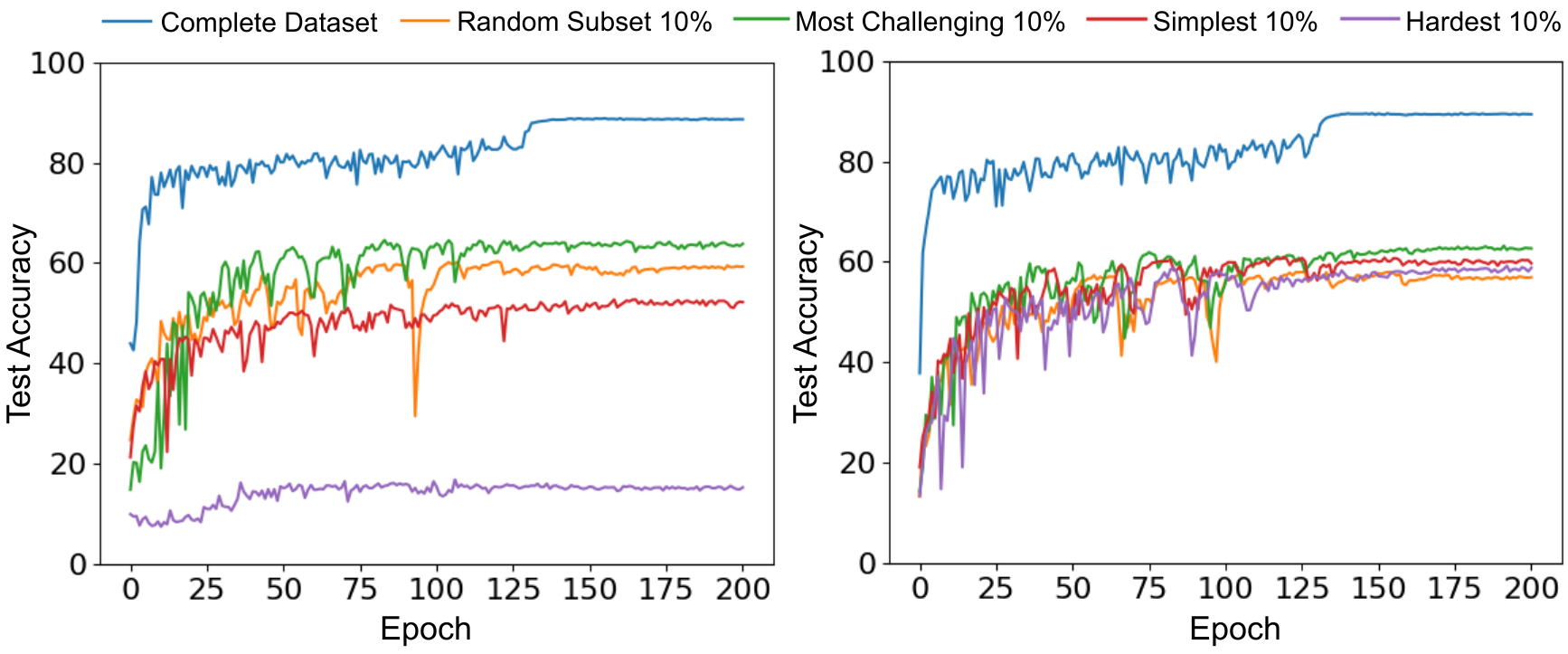}
  \caption{On the left, the figure demonstrates how diverse samples impact the learning process, whereas the right figure clearly portrays the roles that various classes play in facilitating learning.}
  \label{fig:OfflineCarto}
\end{figure}

This study analyzes two scenarios to assess the contributions of classes and samples. In both scenarios, we initially train using the complete dataset. Subsequently, we train on a random subset comprising 10\(\%\) of the dataset. In the first scenario, we focus on the samples' contribution to training. To emulate rehearsal-based methods, which typically retain a small buffer of old classes, we use 10\(\%\) of the dataset. We conduct separate training sessions with the simplest 10\(\%\) of samples, the hardest 10 \(\%\), and the most challenging 10\(\%\). As depicted in Fig.~\ref{fig:SelectedSamples}, the challenging samples appear to be more representative. Furthermore, Fig.~\ref{fig:OfflineCarto} illustrates that training on challenging samples yields superior performance compared to the other types.

In the second scenario, we examine the impact of class contributions. We experiment by allocating a larger number of samples to the simplest class, while randomly selecting 10\(\%\) of the dataset for all classes. This process is repeated for the most challenging and the hardest classes. As shown in Fig.~\ref{fig:OfflineCarto}, allocating more samples to challenging classes leads to higher accuracy.

%% file: sec/6_Experiments.tex
\section{Experiments}
\label{sec:Experiments}

\begin{table*}[t]
\caption{Average End Accuracy (higher is better $\uparrow$).}
\label{Table:AccuracyCIFAR100MiniImageNet}
\setlength{\arrayrulewidth}{0.5pt}
\resizebox{\textwidth}{!}{
\begin{tabular}{c|c|ccccc:c|ccccc:c} \hline
Task   &   & \multicolumn{5}{c:}{Split CIFAR100 \([32 \times 32]\)} & Mean & \multicolumn{5}{c:}{Split Mini-ImageNet \([84 \times 84]\)} & Mean \\ \hline
Buffer &   & \multicolumn{1}{c|}{200}  & \multicolumn{1}{c|}{500}  & \multicolumn{1}{c|}{1000} & \multicolumn{1}{c|}{2000} & \multicolumn{1}{c:}{5000} & & \multicolumn{1}{c|}{200}   & \multicolumn{1}{c|}{500}   & \multicolumn{1}{c|}{1000}  & \multicolumn{1}{c|}{2000}  & 5000  \\ \hline

                           & Base-line   & $7.97_{\pm 0.40}$      & $9.94_{\pm 0.63}$      & $12.07_{\pm 0.49}$   & $14.91_{\pm 0.49}$   & $21.40_{\pm 1.04}$ & 13.26  &  $7.49_{\pm 0.26}$      &  $8.65_{\pm 0.34}$      & $10.44_{\pm 0.24}$    & $13.30_{\pm 0.62}$    & $18.73_{\pm 0.62}$  & 11.72  \\ \cline{2-2}
\multirow{-2}{*}{ER}       & CASP     & \boldmath{$9.14_{\pm 0.38}$}       & \boldmath{$12.35_{\pm 0.64}$}      & \boldmath{$16.14_{\pm 0.48}$}   & \boldmath{$19.89_{\pm 0.56}$}    & \boldmath{$26.51_{\pm 0.64}$} & \textbf{16.81}  & \boldmath{$8.47_{\pm 0.33}$}      & \boldmath{$10.80_{\pm 0.32}$}      & \boldmath{$13.53_{\pm 0.45}$}    & \boldmath{$17.66_{\pm 0.54}$}    & \boldmath{$22.59_{\pm 0.62}$}   & \textbf{14.61}  \\ \hline

                           & Base-line   & $8.03_{\pm 0.42}$      & $9.81_{\pm 0.51}$      & $12.11_{\pm 0.38}$   & $14.59_{\pm 0.53}$   & $20.79_{\pm 0.71}$ & 13.07  & $7.32_{\pm 0.34}$       & $8.36_{\pm 0.34}$       & $9.66_{\pm 0.38}$    & $12.72_{\pm 0.64}$    & $17.65_{\pm 0.71}$  & 11.14 \\ \cline{2-2}
\multirow{-2}{*}{MIR}      & CASP     & \boldmath{$9.35_{\pm 0.26}$}      & \boldmath{$12.37_{\pm 0.52}$}      & \boldmath{$15.78_{\pm 0.50}$}    & \boldmath{$19.86_{\pm 0.58}$}    & \boldmath{$25.87_{\pm 0.97}$}  & \textbf{16.65} & \boldmath{$8.29_{\pm 0.21}$}      & \boldmath{$10.34_{\pm 0.44}$}      & \boldmath{$12.87_{\pm 0.38}$}    & \boldmath{$16.97_{\pm 0.58}$}     & \boldmath{$22.22_{\pm 0.56}$}   & \textbf{14.14}  \\ \hline                  
                           
                           & Base-line    & $17.88_{\pm 0.65}$      & $25.28_{\pm 0.84}$      & $32.58_{\pm 0.63}$   & $39.32_{\pm 0.42}$   & $42.10_{\pm 0.70}$  & 31.43  & $17.37_{\pm 0.64}$       & $26.30_{\pm 0.48}$       & $33.39_{\pm 0.53}$    & $39.53_{\pm 0.47}$    & $42.98_{\pm 0.61}$  & 31.91  \\ \cline{2-2}
\multirow{-2}{*}{SCR}      & CASP      & \boldmath{$20.86_{\pm 0.97}$}    & \boldmath{$27.31_{\pm 0.52}$}    & \boldmath{$33.38_{\pm 0.62}$}    & $39.19_{\pm 0.45}$   & \boldmath{$42.93_{\pm 0.64}$} & \textbf{32.73}  & \boldmath{$21.38_{\pm 0.64}$}    & \boldmath{$29.25_{\pm 0.35}$}      & \boldmath{$34.43_{\pm 0.59}$}    & $38.74_{\pm 0.46}$    & $42.61_{\pm 0.47}$  & \textbf{33.28}  \\ \hline
                           & Base-line    & $17.39_{\pm 0.61}$      &  $23.74_{\pm 0.50}$     & $30.20_{\pm 0.50}$   & $36.68_{\pm 0.54}$   & $44.22_{\pm 0.45}$   & 30.45  & $17.41_{\pm 0.69}$       & $22.38_{\pm 1.02}$       & $28.16_{\pm 0.88}$    & $33.93_{\pm 0.66}$    & $41.14_{\pm 0.84}$  & 28.60  \\ \cline{2-2}
\multirow{-2}{*}{DVC}      & CASP    & \boldmath{$19.32_{\pm 0.79}$}      & \boldmath{$25.56_{\pm 0.71}$}      & \boldmath{$31.50_{\pm 0.63}$}   & \boldmath{$37.97_{\pm 0.67}$}   & \boldmath{$44.58_{\pm 0.70}$}  & \textbf{31.79}  & \boldmath{$20.10_{\pm 0.58}$}       & \boldmath{$25.54_{\pm 0.74}$}      & \boldmath{$30.40_{\pm 0.68}$}    & \boldmath{$35.68_{\pm 0.58}$}    & \boldmath{$42.63_{\pm 0.73}$}  & \textbf{30.87}  \\ \hline
                           & Base-line    & $18.64_{\pm 0.52}$      & $24.82_{\pm 0.78}$      & $31.09_{\pm 0.41}$   & $36.69_{\pm 0.56}$   & $43.89_{\pm 0.64}$  & 31.03  & $16.92_{\pm 0.73}$       & $22.08_{\pm 1.14}$       & $28.66_{\pm 0.61}$    & $34.44_{\pm 1.51}$    & $41.63_{\pm 1.12}$   & 28.75  \\ \cline{2-2}
\multirow{-2}{*}{PCR}      & CASP    & \boldmath{$21.70_{\pm 0.60}$}    & \boldmath{$27.85_{\pm 0.51}$}     & \boldmath{$32.87_{\pm 0.58}$}    & \boldmath{$37.91_{\pm 0.65}$}   & \boldmath{$44.41_{\pm 0.62}$}  & \textbf{32.95}  & \boldmath{$19.95_{\pm 0.84}$}     & \boldmath{$25.07_{\pm 0.89}$}   & \boldmath{$30.56_{\pm 0.70}$}   & \boldmath{$36.39_{\pm 0.34}$}  & \boldmath{$43.08_{\pm 0.91}$}    & \textbf{31.01}   \\ \hline
 \end{tabular}
}
\end{table*}

\begin{table*}[t]
\caption{Average End Forgetting (lower is better $\downarrow$).}
\label{Table:ForgettingCIFAR100MiniImageNet}
\setlength{\arrayrulewidth}{0.5pt}
\resizebox{\textwidth}{!}{
\begin{tabular}{c|c|ccccc:c|ccccc:c} \hline
Task   &  & \multicolumn{5}{c:}{Split CIFAR100 \([32 \times 32]\)} & Mean & \multicolumn{5}{c:}{Split Mini-ImageNet \([84 \times 84]\)} & Mean \\ \hline
Buffer &  & \multicolumn{1}{c|}{200}  & \multicolumn{1}{c|}{500}  & \multicolumn{1}{c|}{1000} & \multicolumn{1}{c|}{2000} & \multicolumn{1}{c:}{5000} &  & \multicolumn{1}{|c}{200}   & \multicolumn{1}{c|}{500}   & \multicolumn{1}{c|}{1000}  & \multicolumn{1}{c|}{2000}  & 5000  \\ \hline

                           & Base-line    & $59.65_{\pm 0.93}$     & $58.05_{\pm 0.83}$      & $54.66_{\pm 0.85}$   & $49.87_{\pm 0.61}$   & $40.59_{\pm 0.80}$  & 52.56  & $59.32_{\pm 0.92}$       & $57.33_{\pm 0.86}$       & $55.62_{\pm 0.52}$    & $51.42_{\pm 1.05}$    & $42.65_{\pm 0.67}$  & 53.27  \\ \cline{2-2}
\multirow{-2}{*}{ER}       & CASP     &  $59.85_{\pm 0.93}$     & \boldmath{$56.35_{\pm 0.90}$}     & \boldmath{$51.96_{\pm 0.68}$}   & \boldmath{$46.33_{\pm 0.71}$}   & \boldmath{$36.85_{\pm 0.44}$}  & \textbf{50.27}  & \boldmath{$58.90_{\pm 0.89}$}       & \boldmath{$56.31_{\pm 0.66}$}       & \boldmath{$53.08_{\pm 0.92}$}    & \boldmath{$48.45_{\pm 0.91}$}    & \boldmath{$39.58_{\pm 0.58}$}   & \textbf{51.26}  \\ \hline

                           & Base-line     & $60.22_{\pm 1.18}$      & $57.73_{\pm 0.80}$      & $54.30_{\pm 0.69}$   & $49.59_{\pm 0.49}$   & $40.74_{\pm 0.48}$  & 52.52   & $59.09_{\pm 0.94}$       & $57.03_{\pm 1.01}$       & $55.43_{\pm 1.26}$    & $52.07_{\pm 0.91}$    & $43.43_{\pm 0.68}$  & 53.41  \\ \cline{2-2}
\multirow{-2}{*}{MIR}      & CASP       & \boldmath{$59.72_{\pm 1.02}$}      & \boldmath{$56.34_{\pm 0.68}$}      & \boldmath{$51.90_{\pm 0.71}$}    & \boldmath{$46.08_{\pm 0.64}$}    & 
\boldmath{$36.28_{\pm 0.54}$}   & \textbf{50.06}   & \boldmath{$58.51_{\pm 0.58}$}      & \boldmath{$55.75_{\pm 0.88}$}      & \boldmath{$53.42_{\pm 0.76}$}     & \boldmath{$48.79_{\pm 0.63}$}     & 
\boldmath{$39.77_{\pm 0.61}$}   & \textbf{51.25}   \\ \hline                                                   
                           
                           & Base-line    & $57.40_{\pm 1.16}$      & $53.00_{\pm 1.10}$      & $43.64_{\pm 0.82}$   & $32.90_{\pm 0.67}$   & $28.09_{\pm 0.70}$   & 43.01   & $38.63_{\pm 1.36}$       & $35.37_{\pm 1.01}$       & $29.95_{\pm 0.99}$    & $24.01_{\pm 0.88}$    & $20.51_{\pm 0.77}$   & 29.69  \\ \cline{2-2}
\multirow{-2}{*}{SCR}      & CASP      & \boldmath{$57.21_{\pm 1.17}$}      & \boldmath{$51.14_{\pm 0.70}$}      & \boldmath{$43.08_{\pm 0.79}$}    & $34.26_{\pm 0.91}$   & \boldmath{$26.80_{\pm 0.64}$}  & \textbf{42.50}  & \boldmath{$33.84_{\pm 1.38}$}     & \boldmath{$30.44_{\pm 1.18}$}      & \boldmath{$26.31_{\pm 0.80}$}    & \boldmath{$22.50_{\pm 0.81}$}    & \boldmath{$18.04_{\pm 0.62}$}  & \textbf{26.23}  \\ \hline
                           & Base-line    & $63.09_{\pm 0.68}$      & $55.59_{\pm 0.86}$      & $48.08_{\pm 0.95}$   & $39.57_{\pm 0.80}$   & $30.33_{\pm 0.80}$  & 47.33  & $53.68_{\pm 1.39}$       & $48.89_{\pm 1.15}$       & $42.97_{\pm 0.75}$    & $36.16_{\pm 1.31}$    & $27.00_{\pm 0.97}$  & 41.74  \\ \cline{2-2}
\multirow{-2}{*}{DVC}      & CASP      & \boldmath{$61.70_{\pm 1.23}$}      & \boldmath{$54.60_{\pm 0.97}$}      & $48.19_{\pm 1.07}$   & $39.71_{\pm 1.26}$   & $30.76_{\pm 0.99}$  & \textbf{46.99}  & \boldmath{$51.89_{\pm 1.33}$}       & \boldmath{$46.93_{\pm 1.50}$}       & $43.67_{\pm 1.19}$    & \boldmath{$35.99_{\pm 0.80}$}    & $27.96_{\pm 0.57}$  & \textbf{41.29}   \\ \hline
                            & Base-line    & $56.82_{\pm 1.73}$      & $47.41_{\pm 2.14}$      & $37.75_{\pm 1.74}$   & $31.17_{\pm 1.49}$   & $17.88_{\pm 1.31}$  & 38.21  & $55.67_{\pm 2.07}$       & $50.39_{\pm 2.78}$       & $41.55_{\pm 1.09}$    & $35.70_{\pm 3.54}$    & $24.99_{\pm 3.12}$  & 41.66 \\ \cline{2-2}
\multirow{-2}{*}{PCR}       & CASP     & \boldmath{$52.37_{\pm 1.18}$}       & \boldmath{$44.26_{\pm 1.65}$}    & \boldmath{$37.26_{\pm 1.07}$}   & \boldmath{$29.79_{\pm 1.33}$}  & $18.07_{\pm 1.30}$   & \textbf{36.35}  & \boldmath{$53.65_{\pm 1.67}$}    & \boldmath{$49.32_{\pm 2.33}$}   & $41.67_{\pm 1.16}$    & \boldmath{$33.85_{\pm 1.45}$}   & \boldmath{$22.93_{\pm 2.68}$}  & \textbf{40.28}  \\ \hline
 \end{tabular}
}
\end{table*}

\begin{figure*}[t]
  \centering
  \includegraphics[width=0.91\textwidth]{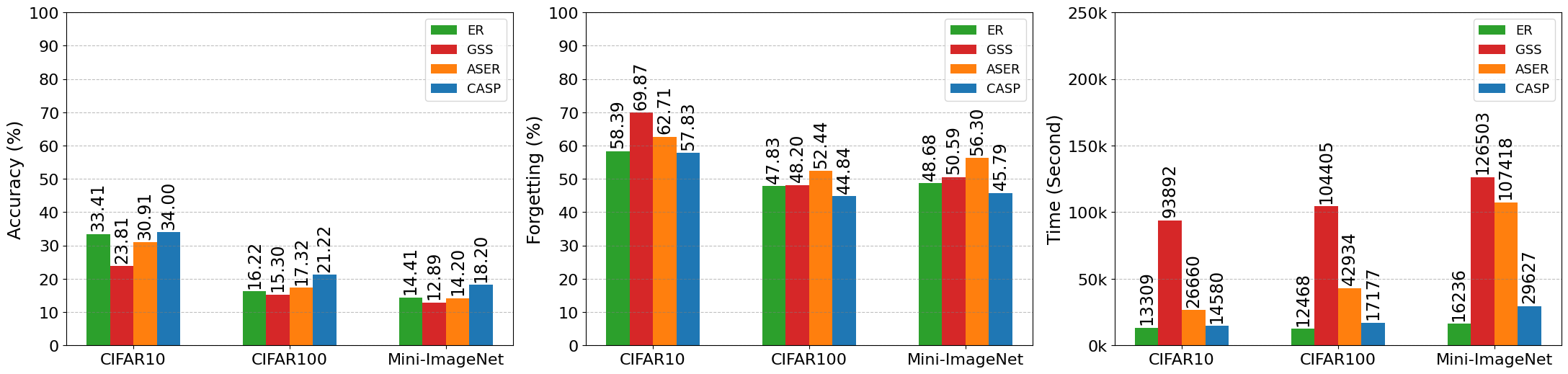}
  \caption{Comparing CASP with two ER-based methods, GSS and ASER. \textit{Left}:Average end accuracy, \textit{Middle}: Average end forgetting, \textit{Right}: Running times across three datasets.}
  \label{fig:updating}
\end{figure*}

\begin{figure*}[t]
  \centering
  \includegraphics[width=0.92\textwidth]{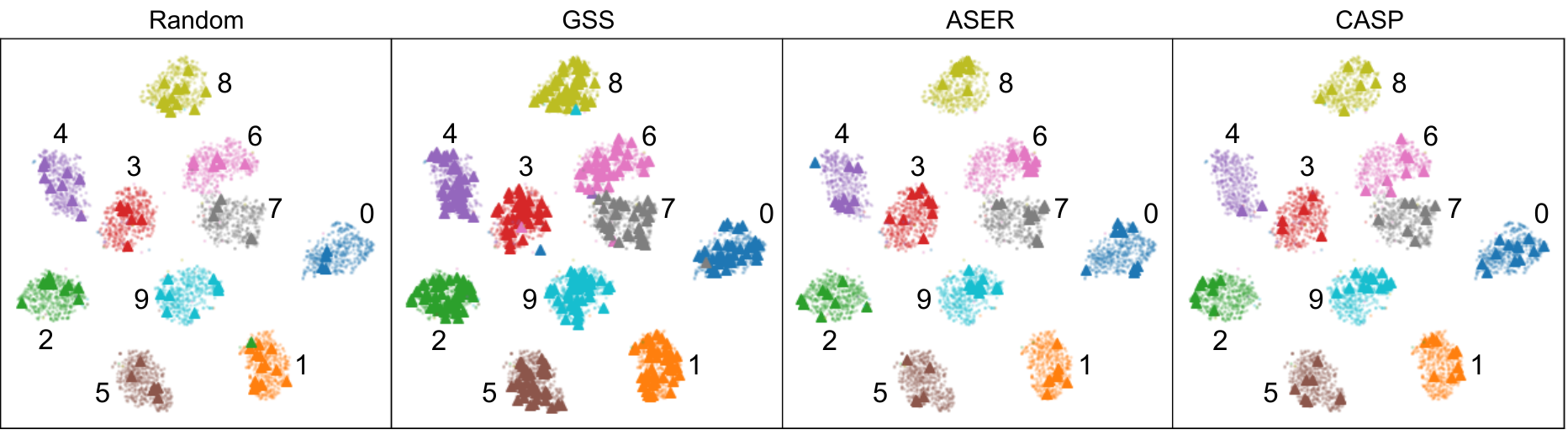}
  \caption{Displaying the embedding space of Task 1 from Split CIFAR100 using 2D t-SNE visualization. Samples from Task 1 that remain in the buffer at the conclusion of Task 10's training are represented by triangles in the figures. The term 'Random' refers to the ER method that lacks a specific policy for memory updates.}
  \label{fig:tsne}
\end{figure*}

\begin{table*}[t]
\caption{Average End Accuracy (higher is better $\uparrow$) for various sample/class strategies on Split CIFAR100 with a buffer size of 1000.}
\label{Table:AblationAccuracy}
\setlength{\arrayrulewidth}{0.5pt}
\centering
\resizebox{0.84\textwidth}{!}{
\begin{tabular}{c|c|c|c|c|c||c|c|c|c||c} \hline
Class Strategy &  & \multicolumn{4}{c||}{Hard} & \multicolumn{4}{c||}{Simple} & \multicolumn{1}{c}{No Policy} \\
\hline
Sample Strategy & & \multicolumn{1}{c|}{Random} & \multicolumn{1}{c|}{Hard} & \multicolumn{1}{c|}{Simple} & \multicolumn{1}{c||}{Challenging} & \multicolumn{1}{c|}{Random} & \multicolumn{1}{c|}{Hard} & \multicolumn{1}{c|}{Simple} & \multicolumn{1}{c||}{Challenging}  & \multicolumn{1}{c}{Random} \\ \hline

                            & i.i.d.    & $31.27_{\pm 1.16}$    & $19.81_{\pm 0.48}$    & $31.38_{\pm 0.50}$                      & $32.21_{\pm 0.63}$    & $31.44_{\pm 0.74}$    & $21.68_{\pm 0.33}$                      & $32.07_{\pm 0.50}$    & $32.71_{\pm 0.53}$      & $31.09_{\pm 0.41}$  \\ \cline{2-2}
                            
     \multirow{-2}{*}{PCR}  & OOD       & $22.50_{\pm 0.48}$    & $14.33_{\pm 0.27}$    & $23.16_{\pm 0.45}$    & $23.47_{\pm 0.60}$     & $22.92_{\pm 0.52}$    & $15.72_{\pm 0.25}$                                 & $23.89_{\pm 0.42}$     & $24.12_{\pm 0.55}$    & $22.32_{\pm 0.26}$   \\ \hline \hline
     
Class Strategy &  & \multicolumn{4}{c||}{Balanced} & \multicolumn{4}{c||}{Challenging} \\ \cline{0-9}
Sample Strategy & & \multicolumn{1}{c|}{Random} & \multicolumn{1}{c|}{Hard} & \multicolumn{1}{c|}{Simple} & \multicolumn{1}{c||}{Challenging} & \multicolumn{1}{c|}{Random} & \multicolumn{1}{c|}{Hard} & \multicolumn{1}{c|}{Simple} & \multicolumn{1}{c||}{Challenging} \\ \cline{0-9}

                            & i.i.d.    & $31.64_{\pm 0.63}$    & $19.79_{\pm 0.44}$    & $31.79_{\pm 0.55}$                      & $31.68_{\pm 0.91}$    & $32.06_{\pm 0.64}$    & $21.65_{\pm 0.63}$                      & $32.07_{\pm 0.72}$    & $32.87_{\pm 0.58}$  \\ \cline{2-2}
                            
     \multirow{-2}{*}{PCR}  & OOD       & $22.72_{\pm 0.32}$    & $14.45_{\pm 0.16}$    & $23.52_{\pm 0.37}$                                             & $23.30_{\pm 0.69}$    & $23.27_{\pm 0.55}$    & $15.71_{\pm 0.43}$                                             & $23.84_{\pm 0.60}$    & $24.36_{\pm 0.65}$    \\

\end{tabular}
}
\end{table*}

\subsection{Setup}

\paragraph{Datasets.} Our experiments were conducted using three datasets. The first is Split CIFAR10 ~\cite{krizhevsky2009learning}, comprising 5 disjoint tasks, with each task encompassing 2 classes. The second dataset is Split CIFAR100 ~\cite{krizhevsky2009learning}, and the third is Split Mini-ImageNet ~\cite{vinyals2016matching}. Both of these latter datasets feature 10 disjoint tasks, each containing 10 classes.

\paragraph{Metrics.} We employ two standard metrics to evaluate the performance of CASP: \textit{average accuracy} and \textit{average forgetting}. The former assesses the overall performance, while the latter measures the extent of knowledge loss that the algorithm experiences over time, as outlined in ~\cite{chaudhry2018riemannian, chaudhry2019tiny}. 

Average Accuracy is defined as follows: Let \(\alpha_{i,j}\) denote the accuracy on the held-out test set for task j after training the network on tasks 1 through i. Once all T tasks have been trained, the Average End Accuracy is calculated using the formula:

\begin{equation}
    \text{Average End Accuracy} = \frac{1}{T}\sum_{j=1}^T\alpha_{T,j}
\end{equation}

Average Forgetting measures the degree of knowledge loss regarding task j after the model has been trained on task i. Once all T tasks have been trained, the Average End Forgetting is calculated using the formula:

\begin{equation}
    \text{Average End Forgetting} = \frac{1}{T-1}\sum_{j=1}^{T-1}\beta_{T,j}
\end{equation}

Here, \(\beta_{i,j}\) is calculated as \( \text{max}(\alpha_{1,j}, \dots, \alpha_{i-1,j}) - \alpha_{i,j}\), representing the reduction in model performance on task j as a result of training on subsequent tasks.

\paragraph{CASP Settings.} In our experiments across all three datasets, we employ a ResNet18 ~\cite{he2016deep} model, training it from scratch for each task. The optimization is carried out using stochastic gradient descent (SGD), configured with a learning rate of 0.1 and momentum of 0.9. We also apply a weight decay of 5e-4. To regulate the learning rate, a cosine annealing scheduler is utilized, which is set for a maximum of 200 iterations. Cross-entropy loss is used as the loss function. Regarding the number of training epochs, we set it to 4 for Split CIFAR10 and 8 for both Split CIFAR100 and Split Mini-ImageNet. The impact of varying epoch numbers on each dataset is further analyzed in our ablation study, as detailed in \ref{CASPEpoch}.

\paragraph{Baseline Settings} We employ a Reduced ResNet18 as the backbone model for all baselines across the three datasets, consistent with previous studies ~\cite{lopez2017gradient, chaudhry2019tiny, NEURIPS2019_15825aee}. Following the experimental framework presented in ~\cite{lin2023pcr}, we conduct 10 runs, setting both the current batch and memory batch sizes to 10, and employ an SGD optimizer with a learning rate of 0.1. For the hyperparameters of MIR, GSS, ASER, and SCR we adhere to the specifications in ~\cite{mai2021supervised}. Meanwhile, we follow ~\cite{gu2022not} for DVC hyperparameters.  In the experiments detailed in Sections \ref{EvaluatingCASP}, and \ref{Ablation} the number of epochs for the CL model is set to 25, whereas for Section \ref{ComparisonGSSASER}, it is set to 10. We examine the effects of varying epoch numbers in the supplementary material. To better replicate real-world scenarios, where class order is not predetermined, we set the 'Fixed Class Order' hyperparameter to 'False'. This adjustment ensures a more realistic simulation of real-world conditions. For additional details, please refer to the supplementary material.

\subsection{Evaluating the CASP effectiveness}
\label{EvaluatingCASP}

The integration of CASP with various random updating methods, namely ER, MIR, SCR, DVC, and PCR, demonstrates its superior effectiveness. These methods, in the process of learning new tasks, employ a uniform sampling strategy to update the buffer with each new batch. However, treating all samples within a batch uniformly inadvertently leads to class imbalances in the stored samples in the buffer. In fact, allocating different numbers of samples for each class is somewhat random. Contrary to previous  methods, CASP not only recognizes but also prioritizes the contributions of each class. It goes a step further by strategically incorporating the most informative samples for every class, thereby optimizing the learning process.

Table \ref{Table:AccuracyCIFAR100MiniImageNet} clearly demonstrates the efficacy of our approach. We evaluated the performance of previous methods with different buffer sizes and also reported their average performance across all sizes. The results show that in most cases (different sizes and average across all buffer sizes), our method efficiently improves performance.  For the Split CIFAR100 dataset, CASP improves the performance of ER by an impressive 3.55\(\%\), MIR by 3.58\(\%\), SCR by 1.30\(\%\), DVC by 1.34\(\%\), and PCR by 1.92\(\%\). Similarly, in the Split Mini-ImageNet context, the enhancement rates were 2.89\(\%\) for ER, 3.00\(\%\) for MIR, 1.37\(\%\) for SCR, 2.27\(\%\) for DVC, and 2.26\(\%\) for PCR. 

Moreover, as detailed in Table \ref{Table:ForgettingCIFAR100MiniImageNet}, CASP substantially mitigated the issue of forgetting. In the Split CIFAR100 dataset, the improvements were 2.29\(\%\) with ER, 
2.46\(\%\) with MIR, 0.51\(\%\) with SCR, 0.34\(\%\) with DVC, and 1.86\(\%\) with PCR. In the case of Split Mini-ImageNet, we observed reductions in forgetting by 2.01\(\%\) with ER, 2.16\(\%\) with MIR, 3.46\(\%\) with SCR, 0.45\(\%\) with DVC, and 1.38\(\%\) with PCR. See the supplementary material for results on CIFAR10.

\subsection{Comparison with GSS and ASER Methods}
\label{ComparisonGSSASER}
To ensure a balanced comparison with other methods implementing a buffer update policy, we focus on two Experience Replay (ER) based approaches, \textbf{GSS} and \textbf{ASER}. Unlike ER, which lacks a specific policy for memory updates and operates on a random basis, GSS and ASER enhance the ER framework by incorporating a strategy for memory policy updates. This distinction is critical for our analysis, as we utilize the ER method as a reference point. By integrating our algorithm with ER (as detailed in Algorithm \ref{alg:er_method}), we effectively compare it against the GSS and ASER methods, highlighting the improvements brought about by our approach.

As depicted in Fig.~\ref{fig:updating}, our method demonstrates superior accuracy over GSS and ASER. For Split CIFAR10, we achieved a 10.19\(\%\) and 3.09\(\%\) improvement over GSS and ASER, respectively, while also enhancing ER's performance by 0.59\(\%\). In the case of Split CIFAR100, our method surpasses GSS and ASER by margins of 5.92\(\%\) and 3.90\(\%\), respectively, and boosts ER by 5.00\(\%\). For Split Mini-ImageNet, we outperform GSS and ASER by 5.31\(\%\) and 4.00\(\%\), respectively, and elevate ER by 3.79\(\%\).

Regarding the aspect of forgetting, as depicted in Fig.~\ref{fig:updating}, our approach continues to outpace both GSS and ASER. In Split CIFAR10, the improvement margins are 12.04\(\%\) and 4.88\(\%\) over GSS and ASER, respectively, with a 0.56\(\%\) enhancement in ER. For Split CIFAR100, the advancements are 3.36\(\%\) and 7.60\(\%\) over GSS and ASER, respectively, alongside a 2.99\(\%\) boost in ER. In the Split Mini-ImageNet scenario, we surpass GSS and ASER by 4.80\(\%\) and 10.51\(\%\), respectively, and improve ER by 2.89\(\%\).

As demonstrated, our method is superior to both GSS and ASER in terms of accuracy and forgetting, but the main drawback of these methods is their higher computational cost compared to ours. The enhanced speed of our approach is due to two key factors. Firstly, we employ the CASP model to rapidly ascertain the contribution of each class and sample by training it with all task data simultaneously for a few epochs. Secondly, our buffer update strategy is more efficient as it operates on a task-by-task basis, unlike the batch-by-batch approach used by GSS and ASER. This efficiency is evident in Fig.~\ref{fig:updating}, where our method shows only a slight increase in computational cost compared to ER, which does not employ a specific updating policy, while significantly outpacing GSS and ASER in terms of speed (For more details, see the supplementary material).


\paragraph{The Underlying Challenges in Amplifying ER with GSS/ASER.} The inability of GSS/ASER to significantly bolster ER stems from two pivotal issues. First and foremost, ASER exhibits a marked susceptibility to over-fitting, a phenomenon extensively documented in the supplementary material. Contrary to expectations, where prolonged training should ideally translate into enhanced model performance, ASER paradoxically shows a regression in effectiveness when trained beyond a single epoch.

Secondly, the choice between fixed and randomized class orders in the realm of CL emerges as a critical juncture. A fixed order, while facilitating a more straightforward learning trajectory with its predictable patterns, often fails to encapsulate the dynamic and unpredictable nature of real-world data. In stark contrast, a randomized order, though it mirrors real-world unpredictability and presents a more formidable challenge to the learning system, simultaneously raises the specter of catastrophic forgetting. Hence, as a non-fixed order is in play, GSS and ASER find themselves at odds with adapting to real-life scenarios, leading to a notable failure in enhancing ER.

\paragraph{Deciphering Buffer Update Strategies: A t-SNE Analysis} To illustrate the specific samples selected by each method from various classes and the number of samples per class for buffer updates, we utilize t-SNE visualization ~\cite{van2008visualizing}. In the context of Split CIFAR100 with a buffer size of 1000, we analyze the samples of task one from the buffer at the conclusion of training across all tasks and show them by triangles in Fig.~\ref{fig:tsne}. For finding the distributions, samples are first processed through a fine-tuned encoder to extract features, which are then input into t-SNE for visualization. The ER method updates the buffer in batches, treating all samples within a current batch or task uniformly. This approach, however, does not account for the individual class contributions, leading to class imbalances. For instance, in class 0, it randomly allocates samples and also covers a specific area of the class's distribution. In contrast, CASP, by considering class contribution, allocates more samples to class 0 and selects boundary samples covering the class's entire distribution based on sample contribution. Apart from ER and CASP, GSS tends to assign numerous samples to the first task, thereby disproportionately focusing on it while neglecting other tasks, leading to task imbalances. GSS also tends to select outlier samples, as seen in the clustering of classes 0, 3, 7, and 8. The figure illustrating ASER demonstrates that this method does not encompass the entire distribution area for most classes, focusing instead on a specific area of the classes' distribution. In addition, ASER also picks outlier samples. But their number is comparatively smaller than that in GSS.

\begin{figure}[t]
  \centering
  \includegraphics[width=0.45\textwidth]{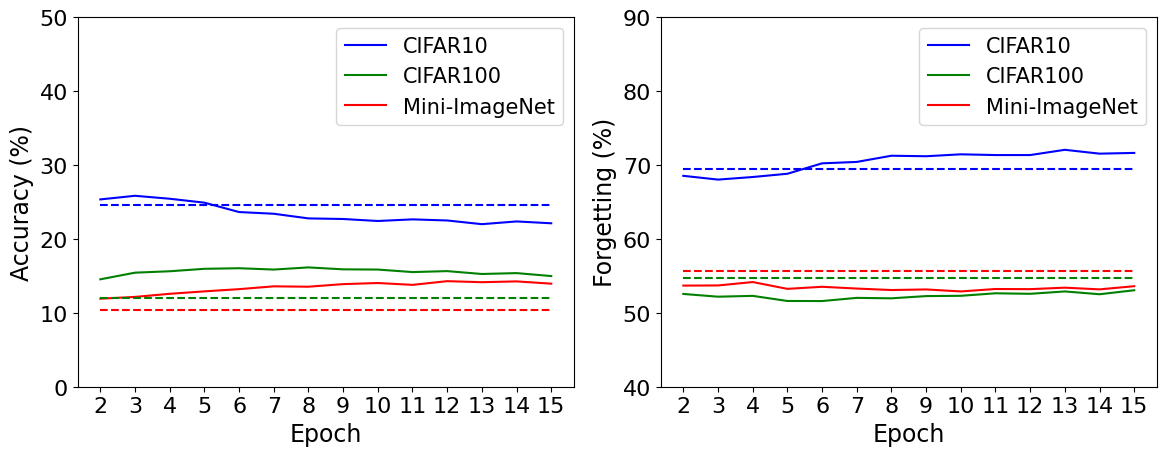}
  \caption{Analyzing the Number of Epochs in the CASP Model: The left figure illustrates the Average End Accuracy, whereas the right figure depicts the Average End Forgetting. Solid lines represent the performance of integrating CASP with Experience Replay (ER), while the dashed lines indicate the baseline performance of the ER method alone.}
  \label{fig:CASPepoch}
\end{figure}

\subsection{Ablation}
\label{Ablation}
\paragraph{Assessing the Impact of Various Class/Sample Strategies on Out-Of-Distribution Generalization}
In Table \ref{Table:AblationAccuracy}, we explore the influence of different Sample/Class Strategies on the generalization capabilities of the learned model. Specifically, our aim is to determine whether the model retains past experiences without compromising its ability to generalize and to identify which approach best maintains the model's Out-Of-Distribution (OOD) generalization.  
For generating OOD images, we adhere to the methodology described in ~\cite{hendrycks2019benchmarking}, which involves applying common corruptions. The data demonstrates that, across each class strategy, selecting challenging samples not only yields superior accuracy on in-distribution samples compared to other sampling strategies but also surpasses them in accuracy on OOD distributions, with the exception of the Balanced Class Strategy. The table further indicates that selecting hard samples reduces accuracy for both in-distribution and OOD samples, as these samples tend to be outliers and are often noisy. Additionally, it is observed that when classes are scored using the Challenging Class Strategy, this approach outperforms other class strategies in terms of accuracy for both in-distribution and OOD samples, except for Hard and Simple Samples under the Simple Class Strategy; however, this exception is not significant. The data also shows that the balanced classes strategy with the Random Sampling Strategy achieves higher accuracy than the baseline (which involves no specific Class Strategy and uniform sample selection, leading to class imbalance), suggesting that disregarding class contributions when creating class imbalances can degrade performance. Finally, the combination of the Challenging Class Strategy with the Challenging Sampling Strategy, known as CASP, outperforms all other strategies in terms of accuracy on both in-distribution and OOD samples. This experiment has also been conducted for SCR and DVC (results are available in the supplementary material).

\paragraph{Impact of Epoch Count on Model Performance and Learning Dynamics}
\label{CASPEpoch}

To demonstrate the impact of the number of epochs in the CASP model on the performance, we evaluated it on Split CIFAR10 (buffer size: 200) and both Split CIFAR100 and Split Mini-ImageNet (buffer size: 1000). This analysis involved comparing the integrated CASP with ER method against ER alone as a baseline. Given its simplicity, Split CIFAR10 requires fewer epochs, as depicted in Fig.~\ref{fig:CASPepoch}. In contrast, the more complex Split CIFAR100 necessitates a higher epoch count, as does Split Mini-ImageNet, which is even more intricate than the other two datasets. The results indicate stable performance across varying epochs, demonstrating CASP's efficiency due to its minimal epoch requirement. Notably, CASP shows enhanced effectiveness on Split CIFAR100 and Split Mini-ImageNet, where complexity is higher due to more challenging and numerous classes.

%% file: sec/7_Conclusion.tex
\section{Conclusion}
\label{sec:Conclusion}

This paper highlights a previously unexplored facet of  CL methodologies. Our findings indicate that uniformly treating all samples and classes in continual learning is suboptimal. Instead, we advocate for a refined strategy focused on retaining the most informative samples from each class. Additionally, we emphasize the importance of paying greater attention to the more challenging classes by preserving a larger number of samples from these classes in the buffer. The data and class selection strategy introduced here improves the performance of current representative methods for CL and also leads to better OOD generalization.

\section*{Acknowledgements} The authors would like to thank Amir Hossein Kargaran and Amirali Molaei.

%% file: sec/8_Supplementary.tex
\clearpage
\setcounter{page}{1}
\maketitlesupplementary

\section{Related Work}
\label{sec:Related-Work}

\paragraph{Data-centric.}Data-centric approaches prioritize data quality over the model. Data-Centric Artificial Intelligence encompasses techniques designed to enhance datasets, enabling the training of models with reduced data requirements ~\cite{motamedi2021data, mazumder2022dataperf}. 

Overlooking the inherent significance of data has led to inaccuracy, bias, and fairness issues in real-world applications ~\cite{mazumder2022dataperf}. Essentially, with high-quality data, we enhance the potential to improve generalization, address bias, and promote safety in data cascades ~\cite{sambasivan2021everyone, aroyo2022data}. 

~\cite{toneva2018empirical, swayamdipta2020dataset} utilize the models' confidence during training to cleanse the dataset. ~\cite{swayamdipta2020dataset} identifies samples with high variability as important for reducing the dataset, whereas ~\cite{toneva2018empirical} deems forgettable samples as crucial for the same purpose. Drawing inspiration from them, we introduce CASP to gauge the importance of classes, alongside samples, during training. Our research reveals that specific classes have a pronounced influence on the training process, a discovery that led us to develop the CASP method.

\paragraph{Randomized Buffer Update Techniques.} \textbf{ER, MIR, SCR, DVC, PCR} ~\cite{chaudhry2019tiny, NEURIPS2019_15825aee, mai2021supervised, gu2022not, lin2023pcr}, update their buffers randomly. \textbf{ER} ~\cite{chaudhry2019tiny} updates memory through reservoir sampling and employs random sampling during memory retrieval. \textbf{MIR} ~\cite{NEURIPS2019_15825aee} focuses on retrieving samples, and is expected to experience increased losses due to the projected updates in model parameters. \textbf{SCR} ~\cite{mai2021supervised} leverages supervised contrastive loss to reduce catastrophic forgetting. \textbf{DVC} ~\cite{gu2022not} focusing on gradient-based sample selection from memory and maximizing mutual information between different views of training images. \textbf{PCR} ~\cite{lin2023pcr} combines proxy-based and contrastive-based replay to mitigate catastrophic forgetting.

\paragraph{Policy-Driven Buffer Update Strategies.} Several methods for updating the buffer have been explored in various studies, including those ~\cite{aljundi2019gradient, shim2021online, de2021continual, jin2021gradient, yoon2021online, tiwari2022gcr, hu2022curiosity, wang2022improving, kumari2022retrospective, yao2023continual}. (\textbf{GSS}) ~\cite{aljundi2019gradient} treats buffer updating as a constrained optimization problem, aiming to maximize the diversity of sample gradients within the memory buffer. \textbf{ASER} ~\cite{shim2021online} employs the Shapley Value (SV) to assign scores to samples based on their propensity for being adversarially proximate to the decision boundaries of emerging classes. (\textbf{CoPE}) ~\cite{de2021continual} utilizes class prototypes and engineers these prototypes to evolve continuously, utilizing a high-momentum-based update mechanism for each observed batch. This method maintains a uniform buffer size across all classes. (\textbf{GMED}) ~\cite{jin2021gradient} refines the buffer content through gradient updates, which are informed by the extent of loss amplification subsequent to a look-ahead update focused exclusively on the new batch. (\textbf{OCS}) ~\cite{yoon2021online} employs a combination of mini-batch gradient similarity and cross-batch diversity to identify the most representative samples. (\textbf{GCR}) ~\cite{tiwari2022gcr} chooses a subset of data that effectively mirrors the gradient profile of the entire dataset encountered up to the present moment, in relation to the current parameters of the model. ~\cite{hu2022curiosity} defines the model's curiosity for individual samples by assessing uncertainty and novelty to identify informative samples that can effectively refine the classification decision boundary. ~\cite{wang2022improving} leverages Distributionally Robust Optimization (DRO) to enhance the complexity of the memory buffer's data distribution over time to cultivate the learning of features with enhanced robustness. (\textbf{RAR}) ~\cite{kumari2022retrospective} generates adversarial examples close to the boundary where forgetting occurs. ~\cite{yao2023continual} dynamically updates memory with discriminating samples to distinguish between old and new classes from the perspective of remedy after forgetting for the first time.

\begin{figure*}[t!]
  \centering
  \includegraphics[width=0.9\textwidth]{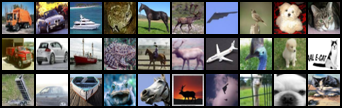}
  \caption{Each column represents a CIFAR10 class. The first row contains 'simple' samples, the second row 'challenging' samples, and the third row 'hard' samples.}
  \label{fig:Samples}
\end{figure*}

\section{Experiments} We conducted our experiments using the NVIDIA GeForce RTX 2080 Ti equipped with CUDA Version 12.1. 

\paragraph{Hyperparameters.} The seed was set to 0. In the MIR approach, we set the number of subsamples to 50. For SCR, we adjusted the temperature for contrastive loss to 0.07 and implemented an MLP for the projection head. In the DVC method, we set the weight of the distance loss to 4.0 for both Split Mini-ImageNet and Split CIFAR100 and 2.0 for Split CIFAR10. For GSS, we determined the number of buffer batches randomly sampled from memory for estimating the maximal gradient cosine similarity score at 10, with a random sampling buffer batch size of 10 for calculating the score. Lastly, for ASER, we set the number of nearest neighbors to 3, utilized mean values of Adversarial SV and Cooperative SV, and established the maximum number of samples per class for random sampling at 3.0 for Split Mini-ImageNet, 2.0 for Split CIFAR100, and 9.0 for Split CIFAR10.

The source codes are available at the following links:

ER, MIR, SCR, GSS, and ASER: \href{https://github.com/RaptorMai/online-continual-learning}{https://github.com/RaptorMai/online-continual-learning}

DVC: \href{https://github.com/YananGu/DVC}{https://github.com/YananGu/DVC}

PCR: \href{https://github.com/FelixHuiweiLin/PCR}{https://github.com/FelixHuiweiLin/PCR}

\begin{figure}[t]
  \includegraphics[width=0.47\textwidth]{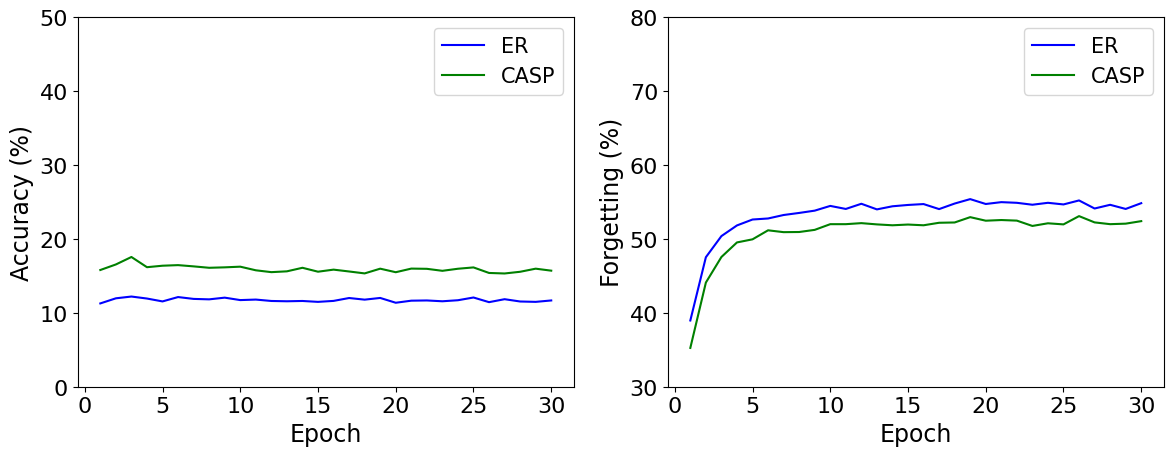}
  \caption{The left figure illustrates the Average End Accuracy, while the right figure depicts the Average End Forgetting. In both figures, the Epoch varies from 1 to 30, corresponding to the main model, not the surrogate model designed specifically for CASP. 'CASP' indicates the integration of CASP with ER, and 'ER' represents ER alone, serving as the baseline. This experiment was conducted using Split CIFAR100 with a buffer size of 1000.}
  \label{fig:ModelEpoch}
\end{figure}

\begin{figure*}[t]
  \centering
  \includegraphics[width=0.9\textwidth]{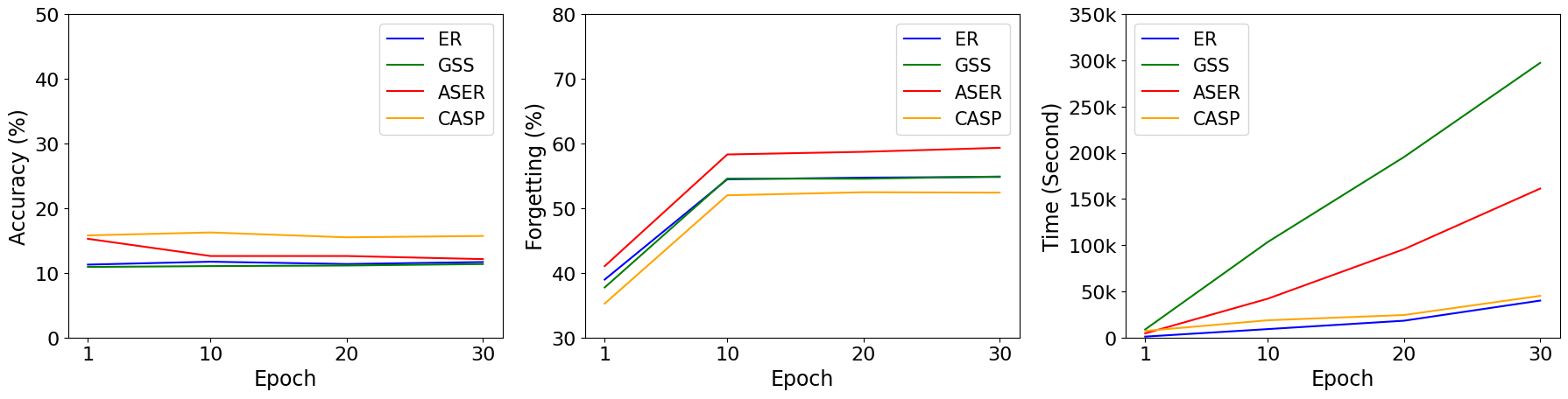}
  \caption{\textit{Left}:Average end accuracy, \textit{Middle}: Average end forgetting, \textit{Right}: Running time. This experiment was conducted using Split CIFAR100 with a buffer size of 1000.}
  \label{fig:UpdatingEpoch}
\end{figure*}

\paragraph{Stability Analysis of CASP Across Various Epochs in CL Model Training.} To demonstrate the stability of CASP throughout different epochs in the training of the CL model, we conducted an analysis across various epoch numbers. For this purpose, CASP was integrated with ER, and the epoch count was varied from 1 to 30. Figure \ref{fig:ModelEpoch} illustrates how changes in the baseline ER correspond to alterations in CASP. This observation leads to the conclusion that CASP maintains stability across a range of epochs.

\paragraph{Comparative Analysis of GSS and ASER Performance Across Epochs} We have also analyzed the performance of GSS and ASER across various epochs in the main model. Specifically, we examined their performance at epochs 1, 10, 20, and 30. The ER model served as our baseline for comparison against CASP, GSS, and ASER. As depicted in Figure \ref{fig:UpdatingEpoch}, it is evident that ASER is prone to overfitting, particularly noticeable in the left figure where an increase in epochs leads to this issue. Conversely, the right figure demonstrates that, with an increase in epoch numbers, the cost escalation for CASP is relatively minor, whereas it is significantly higher for both GSS and ASER.

\begin{table}[ht]
\caption{Average End Accuracy (higher is better $\uparrow$) on the Split CIFAR10 dataset.}
\label{Table:AccuracyCIFAR10}
\setlength{\arrayrulewidth}{0.5pt}
\resizebox{\columnwidth}{!}{
\begin{tabular}{c|c|cccc:c} \hline
Task   &  & \multicolumn{4}{c:}{Split CIFAR10 \([32 \times 32]\)} & Mean  \\ \hline
Buffer &  & \multicolumn{1}{c|}{100}  & \multicolumn{1}{c|}{200}  & \multicolumn{1}{c|}{500}  & \multicolumn{1}{c:}{1000} &   \\ \hline

                           & Base-line  & $21.64_{\pm 1.16}$      & $24.53_{\pm 1.74}$      & $32.53_{\pm 2.47}$      & $40.90_{\pm 2.41}$ & 29.90   \\ \cline{2-2}
\multirow{-2}{*}{ER}       & CASP & \boldmath{$21.78_{\pm 1.25}$}     & \boldmath{$25.43_{\pm 1.12}$}       & $31.89_{\pm 2.22}$      & $38.78_{\pm 2.74}$ & 29.47    \\ \hline

                           & Base-line & $20.99_{\pm 1.21}$      & $23.47_{\pm 0.98}$      & $29.17_{\pm 1.61}$      & $39.17_{\pm 2.59}$  & 28.20   \\ \cline{2-2}
\multirow{-2}{*}{MIR}      & CASP & $20.87_{\pm 1.00}$      & \boldmath{$24.22_{\pm 1.04}$}      & \boldmath{$30.33_{\pm 2.13}$}      & $36.84_{\pm 2.59}$ & 28.07   \\ \hline                  
                           
                           & Base-line & $32.60_{\pm 2.76}$      & $41.06_{\pm 2.84}$      & $57.58_{\pm 2.28}$      & $66.24_{\pm 1.48}$  & 49.37   \\ \cline{2-2}
\multirow{-2}{*}{SCR}      & CASP & \boldmath{$33.51_{\pm 3.13}$}    & \boldmath{$43.02_{\pm 2.36}$}    & $57.10_{\pm 2.21}$      & $65.60_{\pm 1.70}$  & \textbf{49.81}    \\ \hline
                           & Base-line & $25.59_{\pm 1.57}$      & $33.43_{\pm 2.30}$      & $47.32_{\pm 2.42}$      & $57.38_{\pm 1.70}$   & 40.93  \\ \cline{2-2}
\multirow{-2}{*}{DVC}      & CASP & \boldmath{$29.09_{\pm 1.84}$}      & \boldmath{$36.23_{\pm 2.75}$}      & \boldmath{$48.61_{\pm 3.85}$}      & \boldmath{$57.60_{\pm 2.68}$}  & \textbf{42.88}     \\ \hline
                           & Base-line & $48.14_{\pm 1.87}$      & $54.66_{\pm 1.79}$      & $65.07_{\pm 0.75}$      & $70.86_{\pm 1.27}$  & 59.68   \\ \cline{2-2}
\multirow{-2}{*}{PCR}      & CASP & \boldmath{$48.21_{\pm 2.17}$}   & \boldmath{$56.25_{\pm 1.75}$}    & $63.43_{\pm 0.89}$      & $70.38_{\pm 1.10}$  & 59.57     \\ \hline
 \end{tabular}
}
\end{table}

\begin{table}[ht]
\caption{Average End Forgetting (lower is better $\downarrow$) on the Split CIFAR10 dataset.}
\label{Table:ForgettingCIFAR10}
\setlength{\arrayrulewidth}{0.5pt}
\resizebox{\columnwidth}{!}{
\begin{tabular}{c|c|cccc:c} \hline
Task   &  & \multicolumn{4}{c:}{Split CIFAR10 \([32 \times 32]\)} & Mean  \\ \hline
Buffer &  & \multicolumn{1}{c|}{100}  & \multicolumn{1}{c|}{200}  & \multicolumn{1}{c|}{500}  & \multicolumn{1}{c:}{1000}  &   \\ \hline

                           & Base-line  & $72.08_{\pm 0.96}$      & $69.41_{\pm 1.50}$      & $59.81_{\pm 2.27}$      & $50.17_{\pm 2.95}$   & 62.87   \\ \cline{2-2}
\multirow{-2}{*}{ER}       & CASP & \boldmath{$71.70_{\pm 0.78}$}    & \boldmath{$68.37_{\pm 1.06}$}      & $60.47_{\pm 2.36}$      & $52.07_{\pm 3.15}$   & 63.15   \\ \hline

                           & Base-line & $72.90_{\pm 1.04}$      & $70.92_{\pm 1.12}$      & $63.90_{\pm 1.67}$      & $52.56_{\pm 2.86}$    & 65.07    \\ \cline{2-2}
\multirow{-2}{*}{MIR}      & CASP & $73.17_{\pm 1.01}$      & \boldmath{$70.25_{\pm 1.12}$}      & \boldmath{$62.85_{\pm 2.38}$}      & $55.10_{\pm 2.86}$   & 65.34    \\ \hline                                                   
                           
                           & Base-line & $63.73_{\pm 3.28}$      & $55.35_{\pm 3.10}$      & $38.02_{\pm 2.60}$      & $28.92_{\pm 1.67}$   & 46.51   \\ \cline{2-2}
\multirow{-2}{*}{SCR}      & CASP & \boldmath{$62.86_{\pm 3.54}$}      & \boldmath{$53.43_{\pm 2.63}$}     & $38.50_{\pm 2.59}$      & $29.57_{\pm 1.96}$   & \textbf{46.09}   \\ \hline
                           & Base-line & $69.65_{\pm 1.71}$      & $62.27_{\pm 2.49}$      & $47.51_{\pm 2.38}$      & $36.63_{\pm 1.84}$   & 54.02   \\ \cline{2-2}
\multirow{-2}{*}{DVC}      & CASP & \boldmath{$66.03_{\pm 2.19}$}      & \boldmath{$59.27_{\pm 3.00}$}     & \boldmath{$45.99_{\pm 3.95}$}      & \boldmath{$36.32_{\pm 3.12}$}  & \textbf{51.90}   \\ \hline
                            & Base-line & $40.32_{\pm 2.78}$      & $34.23_{\pm 2.62}$      & $22.40_{\pm 1.98}$      & $16.41_{\pm 1.94}$   & 28.34   \\ \cline{2-2}
\multirow{-2}{*}{PCR}       & CASP & $41.85_{\pm 2.40}$      & \boldmath{$33.05_{\pm 2.64}$}    & $25.06_{\pm 1.23}$      & $17.16_{\pm 1.54}$   & 29.28   \\ \hline
 \end{tabular}
}
\end{table}

\begin{table*}[t]
\caption{"Accuracy" denotes the average end accuracy, "Forgetting" signifies the average end forgetting, and "Time" indicates the running time in seconds. The values shown in Figure \ref{fig:updating} represent the data from the 'Mean' columns.}
\label{Table:Updating}
\setlength{\arrayrulewidth}{0.5pt}
\resizebox{\textwidth}{!}{
\begin{tabular}{c|c|ccc:c|ccc:c|ccc:c} \hline

Task   &            & \multicolumn{3}{c:}{Split CIFAR10 \([32 \times 32]\)} & Mean & \multicolumn{3}{c:}{Split CIFAR100 \([32 \times 32]\)} & Mean & \multicolumn{3}{c:}{Split Mini-ImageNet \([84 \times 84]\)} & Mean \\ \hline 
Buffer &            & 200     & 500     & 1000    &      & 1000     & 2000    & 5000    &      & 1000      & 2000      & 5000      &      \\ \hline 
       & Accuracy   & $24.48_{\pm 1.44}$        & $34.11_{\pm 2.25}$        & $41.65_{\pm 1.86}$        & 33.41     & $11.72_{\pm 0.61}$        & $15.18_{\pm 0.96}$        & $21.77_{\pm 0.98}$        & 16.22     & $10.19_{\pm 0.49}$          & $13.54_{\pm 0.38}$          & $19.51_{\pm 0.85}$          & 14.41     \\ \cline{2-2}
ER     & Forgetting & $68.70_{\pm 2.15}$        & $57.63_{\pm 2.57}$        & $48.84_{\pm 1.78}$        & 58.39     & $54.46_{\pm 1.04}$         & $49.25_{\pm 0.92}$        & $39.78_{\pm 0.46}$        & 47.83     & $54.84_{\pm 0.96}$          & $50.27_{\pm 0.95}$          & $40.94_{\pm 0.64}$          & 48.68     \\ \cline{2-2}
       & Time       & 13577 Sec.       & 13234 Sec.       & 13115 Sec.       & 13309 Sec.    & 13124 Sec.        & 12895 Sec.       & 11386 Sec.       & 12468 Sec.    & 16165 Sec.         & 16080 Sec.       & 16236 Sec.        & 16160 Sec.    \\ \hline 
       & Accuracy   & $20.28_{\pm 0.97}$        & $23.16_{\pm 1.47}$        & $27.99_{\pm 1.58}$        & 23.81     & $11.04_{\pm 0.45}$         & $14.43_{\pm 0.78}$        & $20.42_{\pm 0.62}$        & 15.30     & $9.42_{\pm 0.65}$          & $11.94_{\pm 0.70}$          & $17.31_{\pm 0.76}$          & 12.89     \\ \cline{2-2}
GSS    & Forgetting & $74.36_{\pm 1.19}$        & $70.33_{\pm 1.91}$        & $64.92_{\pm 1.79}$        & 69.87     & $54.57_{\pm 1.32}$         & $49.46_{\pm 0.85}$        & $40.58_{\pm 0.72}$        & 48.20     & $55.88_{\pm 0.94}$          & $52.38_{\pm 0.95}$          & $43.51_{\pm 0.71}$          & 50.59     \\ \cline{2-2}
       & Time       & 77755 Sec.       & 101368 Sec.        & 102553 Sec.        & 93892 Sec.    & 103360 Sec.        & 104849 Sec.       & 105005 Sec.       & 104405 Sec.    & 125520 Sec.          & 126356 Sec.          & 127634 Sec.          & 126503 Sec.    \\ \hline 
       & Accuracy   & $22.29_{\pm 0.96}$        & $31.46_{\pm 2.17}$        & $38.99_{\pm 2.40}$        & 30.91     & $12.60_{\pm 0.55}$         & $16.68_{\pm 0.62}$        & $22.69_{\pm 0.90}$        & 17.32     & $11.00_{\pm 0.57}$           & $13.69_{\pm 0.44}$         & $17.92_{\pm 0.46}$        & 14.20   \\ \cline{2-2}
ASER   & Forgetting & $71.80_{\pm 1.46}$        & $62.41_{\pm 2.47}$        & $53.91_{\pm 2.86}$        & 62.71     & $58.30_{\pm 0.87}$         & $52.73_{\pm 0.57}$        & $46.28_{\pm 0.81}$        & 52.44     & $59.86_{\pm 0.66}$          & $56.53_{\pm 0.73}$          & $52.51_{\pm 0.78}$          & 56.30     \\ \cline{2-2}
       & Time       & 26634 Sec.        & 26505 Sec.        & 26842 Sec.        & 26660 Sec.     & 42004 Sec.        & 42707 Sec.        & 44090 Sec.        & 42934 Sec.     & 106897 Sec.    & 105075 Sec.      & 110282 Sec.    & 107418 Sec.  \\ \hline 
       & Accuracy   & \boldmath{\underline{$25.36_{\pm 1.26}$}}       & \boldmath{$33.70_{\pm 2.54}$}        & \boldmath{\underline{$42.95_{\pm 2.06}$}}         & \textbf{\underline{34.00}}     & \boldmath{\underline{$16.24_{\pm 0.75}$}}         & \boldmath{\underline{$20.76_{\pm 0.72}$}}        & \boldmath{\underline{$26.66_{\pm 0.77}$}}        & \textbf{\underline{21.22}}     & \boldmath{\underline{$13.60_{\pm 0.56}$}}        & \boldmath{\underline{$17.65_{\pm 0.53}$}}        & \boldmath{\underline{$23.35_{\pm 0.72}$}}          & \textbf{\underline{18.20}}     \\ \cline{2-2}
CASP   & Forgetting &  \boldmath{\underline{$68.22_{\pm 1.69}$}}        & \boldmath{$58.23_{\pm 2.53}$}        & \boldmath{\underline{$47.03_{\pm 2.47}$}}         & \textbf{\underline{57.83}}     & \boldmath{\underline{$51.99_{\pm 0.92}$}}       & \boldmath{\underline{$45.59_{\pm 1.06}$}}         & \boldmath{\underline{$36.95_{\pm 1.04}$}}        & \textbf{\underline{44.84}}    & \boldmath{\underline{$52.20_{\pm 1.16}$}}         & \boldmath{\underline{$47.21_{\pm 0.88}$}}        & \boldmath{\underline{$37.96_{\pm 0.50}$}}           & \textbf{\underline{45.79}}      \\ \cline{2-2}
       & Time       & \textbf{14632 Sec.}        & \textbf{14510 Sec.}        & \textbf{14597 Sec.}        & \textbf{14580 Sec.}     & \textbf{18053 Sec.}        & \textbf{16589 Sec.}         & \textbf{16889 Sec.}        & \textbf{17177 Sec.}      & \textbf{29588 Sec.}        & \textbf{29589 Sec.}      & \textbf{29705 Sec.}          & \textbf{29627 Sec.}     \\       
\hline
\end{tabular}
}
\end{table*}

\begin{table*}[t]
\caption{Average End Accuracy (higher is better $\uparrow$) across different sample/class strategies in Split CIFAR100, utilizing a buffer size of 1000, for SCR, DVC, and PCR methods.}
\label{Table:AblationAccuracySup}
\setlength{\arrayrulewidth}{0.5pt}
\resizebox{\textwidth}{!}{
\begin{tabular}{c|c|c|c|c|c||c|c|c|c||c} \hline
Class Strategy &  & \multicolumn{4}{c||}{Hard} & \multicolumn{4}{c||}{Simple} & \multicolumn{1}{c}{No Policy} \\
\hline
Sample Strategy & & \multicolumn{1}{c|}{Random} & \multicolumn{1}{c|}{Hard} & \multicolumn{1}{c|}{Simple} & \multicolumn{1}{c||}{Challenging} & \multicolumn{1}{c|}{Random} & \multicolumn{1}{c|}{Hard} & \multicolumn{1}{c|}{Simple} & \multicolumn{1}{c||}{Challenging}  & \multicolumn{1}{c}{Random} \\ \hline

                            & i.i.d.    & $32.08_{\pm 0.34}$    & $20.13_{\pm 0.64}$   & $32.56_{\pm 0.59}$                       & $32.64_{\pm 0.73}$    & $32.59_{\pm 0.67}$   & $21.27_{\pm 0.55}$                       & $32.60_{\pm 0.39}$    & $33.48_{\pm 0.62}$   & $32.58_{\pm 0.63}$   \\ \cline{2-2}
                            
     \multirow{-2}{*}{SCR}  & OOD       & $20.18_{\pm 0.33}$    & $12.66_{\pm 0.39}$   & $21.50_{\pm 0.53}$                                              & $21.23_{\pm 0.68}$    & $21.06_{\pm 0.72}$   & $13.51_{\pm 0.29}$                                              & $21.50_{\pm 0.57}$    & $22.20_{\pm 0.74}$   & $20.55_{\pm 0.52}$   \\ \hline
     
                            & i.i.d.    & $29.96_{\pm 0.90}$    & $19.16_{\pm 0.76}$    & $30.27_{\pm 0.63}$                      & $30.27_{\pm 0.52}$    & $31.08_{\pm 0.56}$    & $20.72_{\pm 0.75}$                      & $30.54_{\pm 0.80}$    & $31.19_{\pm 0.47}$    & $30.20_{\pm 0.50}$  \\ \cline{2-2}
                            
     \multirow{-2}{*}{DVC}  & OOD       & $20.80_{\pm 0.77}$    & $13.22_{\pm 0.31}$    & $21.42_{\pm 0.50}$                                             & $21.15_{\pm 0.36}$    & $21.77_{\pm 0.65}$    & $14.41_{\pm 0.39}$                                             & $21.58_{\pm 0.72}$    & $21.80_{\pm 0.58}$    & $20.94_{\pm 0.58}$  \\ \hline
     
                            & i.i.d.    & $31.27_{\pm 1.16}$    & $19.81_{\pm 0.48}$    & $31.38_{\pm 0.50}$                      & $32.21_{\pm 0.63}$    & $31.44_{\pm 0.74}$    & $21.68_{\pm 0.33}$                      & $32.07_{\pm 0.50}$    & $32.71_{\pm 0.53}$    & $31.09_{\pm 0.41}$  \\ \cline{2-2}
                            
     \multirow{-2}{*}{PCR}  & OOD       & $22.50_{\pm 0.48}$    & $14.33_{\pm 0.27}$    & $23.16_{\pm 0.45}$                                             & $23.47_{\pm 0.60}$    & $22.92_{\pm 0.52}$    & $15.72_{\pm 0.25}$                                             & $23.89_{\pm 0.42}$    & $24.12_{\pm 0.55}$    & $22.32_{\pm 0.26}$ \\ \hline \hline
     
Class Strategy &  & \multicolumn{4}{c||}{Balanced} & \multicolumn{4}{c||}{Challenging} \\ \cline{0-9}
Sample Strategy & & \multicolumn{1}{c|}{Random} & \multicolumn{1}{c|}{Hard} & \multicolumn{1}{c|}{Simple} & \multicolumn{1}{c||}{Challenging} & \multicolumn{1}{c|}{Random} & \multicolumn{1}{c|}{Hard} & \multicolumn{1}{c|}{Simple} & \multicolumn{1}{c||}{Challenging} \\ \cline{0-9}

                            & i.i.d.    & $32.58_{\pm 0.48}$   & $20.75_{\pm 0.52}$    & $32.90_{\pm 0.49}$                       & $33.49_{\pm 0.54}$   & $32.39_{\pm 0.61}$    & $21.58_{\pm 0.26}$                       & $32.73_{\pm 0.57}$   & $33.38_{\pm 0.62}$   \\ \cline{2-2}
                            
     \multirow{-2}{*}{SCR}  & OOD       & $20.38_{\pm 0.31}$  & $13.09_{\pm 0.34}$    & $21.56_{\pm 0.55}$                                               & $21.92_{\pm 0.64}$  & $20.82_{\pm 0.57}$    & $13.77_{\pm 0.37}$                                               & $21.77_{\pm 0.50}$  & $22.04_{\pm 0.50}$   \\ \cline{0-9}
     
                            & i.i.d.    & $30.25_{\pm 0.51}$   & $19.81_{\pm 0.63}$    & $30.41_{\pm 0.44}$                       & $31.34_{\pm 0.54}$   & $30.63_{\pm 0.61}$    & $21.18_{\pm 0.78}$                       & $30.38_{\pm 0.57}$   & $31.50_{\pm 0.63}$  \\ \cline{2-2}
                            
     \multirow{-2}{*}{DVC}  & OOD       & $20.84_{\pm 0.52}$    & $13.77_{\pm 0.42}$    & $21.53_{\pm 0.56}$                                             & $21.92_{\pm 0.57}$    & $21.44_{\pm 0.86}$    & $14.52_{\pm 0.36}$                                             & $21.69_{\pm 0.52}$    & $22.20_{\pm 0.65}$   \\ \cline{0-9}
     
                            & i.i.d.    & $31.64_{\pm 0.63}$    & $19.79_{\pm 0.44}$    & $31.79_{\pm 0.55}$                      & $31.68_{\pm 0.91}$    & $32.06_{\pm 0.64}$    & $21.65_{\pm 0.63}$                      & $32.07_{\pm 0.72}$    & $32.87_{\pm 0.58}$  \\ \cline{2-2}
                            
     \multirow{-2}{*}{PCR}  & OOD       & $22.72_{\pm 0.32}$    & $14.45_{\pm 0.16}$    & $23.52_{\pm 0.37}$                                             & $23.30_{\pm 0.69}$    & $23.27_{\pm 0.55}$    & $15.71_{\pm 0.43}$                                             & $23.84_{\pm 0.60}$    & $24.36_{\pm 0.65}$    \\

\end{tabular}
}
\end{table*}